\documentclass[lettersize,journal]{IEEEtran}
\usepackage{amsmath,amsfonts}
\usepackage{algorithmic}

\usepackage{array}
\usepackage[caption=false,font=normalsize,labelfont=sf,textfont=sf]{subfig}
\usepackage{textcomp}
\usepackage{stfloats}
\usepackage{url}
\usepackage{verbatim}
\usepackage{graphicx}
\usepackage{cite}
\usepackage{hyperref}

\usepackage{pifont}

\usepackage{booktabs} % 导入booktabs包
\usepackage{multirow} % 导入multirow包

\usepackage[mathscr]{euscript}
\usepackage{xcolor}  % Include this for color support

\usepackage[ruled,linesnumbered]{algorithm2e} % 顶/底横线 + 行号（单栏浮动体）
\DontPrintSemicolon
\SetKwInput{KwIn}{Input}
\SetKwInput{KwOut}{Output}
\SetKwFor{While}{\textbf{while}}{\textbf{do}}{\textbf{end while}}

\begin{document}
\title{Language-Guided Long Horizon Manipulation with LLM-based Planning and Visual Perception}

\author{
Changshi Zhou, Haichuan Xu, Ningquan Gu, Zhipeng Wang, Bin Cheng, Pengpeng Zhang,\\
Yanchao Dong, Mitsuhiro Hayashibe, Yanmin Zhou, Bin~He
}

\maketitle

\begin{abstract}
Language-guided long-horizon manipulation of deformable objects presents significant challenges due to high degrees of freedom, complex dynamics, and the need for accurate vision-language grounding. In this work, we focus on multi-step cloth folding—a representative deformable-object manipulation task—requiring both structured long-horizon planning and fine-grained visual perception. To this end, we propose a unified framework that integrates a Large Language Model (LLM)-based planner, a Vision Language Model (VLM)-based perception system, and a task execution module. Specifically, The LLM-based planner decomposes high-level language instructions into low-level action primitives, bridging the semantic–execution gap, aligning perception with action, and enhancing generalization. The VLM-based perception module employs a SigLIP2-driven architecture with a novel bidirectional cross-attention fusion mechanism and Weight-Decomposed Low-Rank Adaptation (DoRA)-based fine-tuning to achieve language-conditioned fine-grained visual grounding. Experiments in both simulation and real-world settings demonstrate the method’s effectiveness. In simulation, it outperforms state-of-the-art (SOTA) baselines, achieving improvements of 2.23\%, 1.87\%, and 33.3\% on seen instructions, unseen instructions, and unseen tasks, respectively. On a real robot, it robustly executes multi-step folding sequences from language instructions across diverse cloth materials and configurations, demonstrating strong generalization in practical scenarios. Project page: \href{https://language-guided.netlify.app/}{https://language-guided.netlify.app/}.
\end{abstract}


\section{Introduction}
\label{sec:introduction}
\IEEEPARstart{E}{nabling} robots to follow natural language commands \cite{mon2025embodied} in the physical world is a long-standing goal in robotics and embodied AI. Among the many domains where this capability is desired, cloth manipulation \cite{9097275} presents particularly formidable challenges. Unlike rigid objects \cite{10611491} \cite{gu2025tactilealoha}, cloth exhibits high degrees of freedom, self-occlusion, and shape variability, making perception and control significantly more difficult. At the same time, folding clothes \cite{wu2024unigarmentmanip} is a common real-world task in domestic and industrial settings, where seamless human-robot interaction through language instructions can greatly improve usability. However, how to translate high-level linguistic goals into low-level manipulation for deformable objects remains largely unexplored.

In recent years, driven by the success of vision-language foundation models (VLMs) \cite{diao2025soundmind,encoder}, many works have explored their use in language-guided robotic manipulation. For example, CLIPort \cite{pmlr-v164-shridhar22a} employs CLIP to map language goals to visual regions, enabling open-vocabulary manipulation through dense correspondence. RT-2 \cite{brohan2023rt} and RoboFlamingo \cite{li2024vision} extend this idea by training vision-language policies over large datasets of robotic trajectories. Deng et al. \cite{deng2024generalizable} and Barbany et al.  \cite{barbany2025bifold} further applied this paradigm to cloth manipulation tasks, using language instructions to guide robot actions. While these models demonstrate promising zero-shot generalization in a variety of manipulation tasks, they remain limited in handling deformable materials and executing temporally extended instructions.  In particular, they frequently struggle to interpret high-level commands involving multi-step cloth folding instructions, such as ``\textit{Fold the T-Shirt in thirds''}. 

At the same time, the application of large language models (LLMs) in robot planning, such as LiP-LLM \cite{obata2024lip}, has shown strong capability in decomposing long-horizon tasks into atomic actions or sub-goals. These methods leverage the symbolic reasoning strength of LLMs to generate action plans aligned with human intent. However, most are validated only in rigid-object domains, and their generalizability to tasks involving high visual ambiguity and physical deformability, like cloth folding, remains underexplored.  Therefore, how to effectively combine the reasoning abilities of LLMs \cite{jiang2025hiddendetect, bi2024llava} with the visual grounding capabilities of VLMs \cite{diao2025temporal, FineCIR} to improve the generalization and interpretability of cloth manipulation is still an open challenge.

To address this, we propose a unified framework that integrates speech-based task input, LLM-driven task planning, and VLM-based multimodal perception (see Figure 1). Specifically, the process begins with a spoken command, which is transcribed into text by an Automatic Speech Recognition (ASR) module and then decomposed into interpretable sub-tasks by an LLM-based planner. For instance, a high-level command such as ``\textit{Fold the sleeve towards the inner of the T-Shirt}'' is decomposed into two precise sub-tasks: \textit{(1) Grasp the left sleeve of the T-Shirt and place it to the left middle part}; \textit{(2) Grasp the right sleeve of the T-Shirt and place it to the right middle part}.

\begin{figure*}[t]
\centering
\includegraphics[width=1\linewidth]{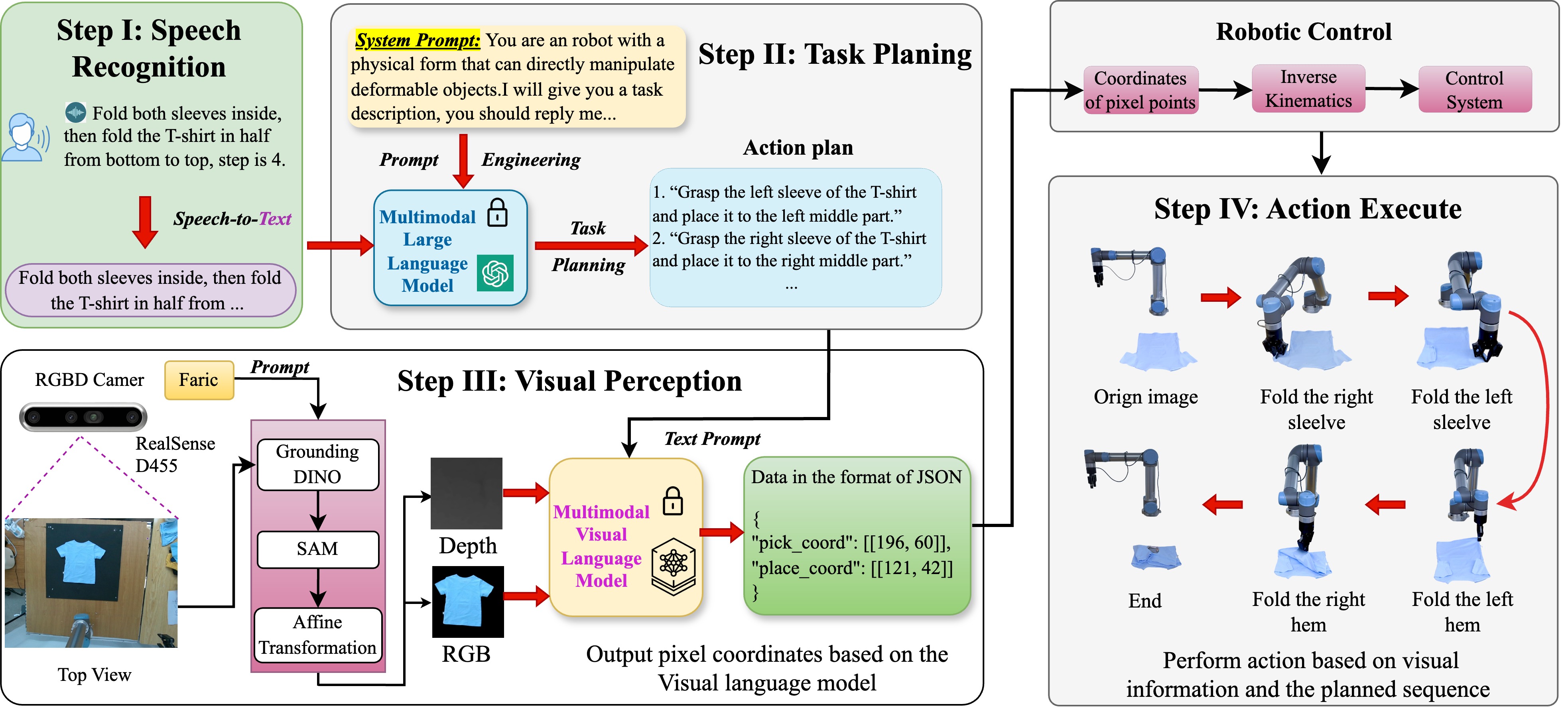}
\caption{\textbf{Method Overview.} An illustration of the robotic-arm embodied LLM system in the real world, showcasing the integrated workflow of Automatic Speech Recognition, Task Planning, Visual Perception, and Action Execution in a cloth manipulation task.}
\end{figure*}

On the perception side, we leverage a frozen SigLIP2 encoder \cite{tschannen2025siglip}, pretrained on large-scale image-text pairs, to extract rich semantic features from RGB-D observations and task instructions. To align these modalities, we design a bidirectional cross-attention fusion module that enables sub-task instructions to focus on relevant visual regions while allowing visual features to be conditioned on task semantics. Furthermore, we employ Weight-Decomposed Low-Rank Adaptation (DoRA) \cite{liu2024dora} to efficiently fine-tune the model for fabric-specific tasks, enhancing its sensitivity to fine-grained cloth features. We evaluate the proposed method in both simulation and real-world settings. In simulation, it surpasses state-of-the-art baselines \cite{barbany2025bifold} on multi-task cloth folding benchmarks. On a real UR5 robot, it generates reliable long-horizon task plans and predicts accurate end-effector poses for each atomic action, demonstrating robustness across diverse cloth configurations. In summary, our contributions are as follows:

\begin{itemize}
\item We propose a unified framework for language-guided cloth manipulation that integrates speech-based input, LLM-driven task planning, and VLM-based visual grounding, supporting full-stack perception-to-action control from language instructions.

\item We introduce an LLM-based planner that decomposes high-level instructions into interpretable sub-tasks aligned with robot-executable primitives, improving the structure and reliability of long-horizon cloth folding tasks.

\item We design a bidirectional cross-attention fusion mechanism that aligns textual and visual representations, built on a frozen SigLIP2 encoder and adapted via DoRA for fine-grained cloth-specific perception.

\item We demonstrate, through extensive simulation and real-world experiments, that the proposed method effectively executes language-guided cloth manipulation tasks and generalizes across diverse fabric types and configurations.

\end{itemize}

\section{Related work}
\subsection{VLMs for Robotics}
VLMs have enabled robots to associate language with visual observations, supporting open-vocabulary and multitask manipulation. Representative works include MmAP \cite{xin2024mmap} and CLIPort \cite{pmlr-v164-shridhar22a}, which align language goals with image regions for pick-and-place actions. More recent models like Gaussiangrasper \cite{zheng2024gaussiangrasper}, Tinyvla \cite{wen2025tinyvla}, and Naturalvlm \cite{xu2024naturalvlm} combine vision-language inputs with large-scale policy learning for generalizable control. While effective in grounding low-level instructions to visual input, these models typically treat language as static goal descriptors and lack the semantic reasoning needed to interpret complex, multi-step commands such as ``\textit{Fold the T-Shirt in thirds''}. This limits their ability to perform high-level task abstraction and decision-making. In our work, we address this gap by pairing a VLM-based perception module with an LLM-based planner, where the latter decomposes high-level instructions into interpretable sub-tasks, and the former grounds each sub-task.

\subsection{LLMs for Robotics}
LLMs\cite{jiang2025screencoder,bi2025prism} have recently shown strong capabilities in robotic task planning by leveraging their compositional reasoning to translate high-level goals into structured sub-tasks. Zhao et al. \cite{zhao2025learning} and Li et al. \cite{li2025language} use prompting or code generation to convert natural language into executable actions or policy scripts, allowing robots to perform long-horizon tasks through structured task decomposition. Chain-of-Thought prompting \cite{wei2022chain} further improves decomposition by guiding LLMs through explicit intermediate reasoning steps, while methods like Text2Motion \cite{lin2023text2motion} and multi-task instruction-following models \cite{wang2023learning} demonstrate generalization across diverse robot behaviors. Despite these advances, most LLM-based approaches are evaluated in rigid-object domains with symbolic or low-dimensional state spaces, where perception is relatively simple and object dynamics are stable. Their effectiveness in visually complex, deformable manipulation tasks remains underexplored. In this work, we address this gap by leveraging LLMs for task-level decomposition from a commands and pairing them with a VLM-based perception module for grounding each sub-task in high-dimensional, ambiguous visual scenes.

\begin{algorithm}[t]
\caption{Language-driven Task Planning and Action Execution}
\KwIn{User command $S$; initial observation $o_t \in \mathbb{R}^{C\times H\times W}$;
system prompt $p$; Automatic Speech Recognition module $f_{\mathrm{asr}}$;
task planner $\mathcal{M}_T$; visual module $\mathcal{M}_V$; action executor $\mathcal{M}_A$.}
\KwOut{Task executed by following the parsed action sequence.}

Transcribe speech: $L_q \gets f_{\mathrm{asr}}(S)$\;
Parse instruction: $L = (\ell_1,\ldots,\ell_n) \gets \mathcal{M}_T(p, L_q)$\;
Initialize: $t \gets 1$\;

\While{$t \le n$}{
    Predict action: $a_t \gets \mathcal{M}_V(o_t,\ell_t)$\;
    Execute action: $\mathcal{M}_A(a_t)$\;
    Update observation: $o_{t+1} \gets \mathcal{M}_V.\mathrm{observe}()$\;
    $t \gets t + 1$\;
}
\end{algorithm}

\subsection{Cloth Manipulation}
Cloth manipulation remains a long-standing challenge in robotics due to the complexity of fabric dynamics, including self-occlusion \cite{huang2022mesh}, degrees of freedom \cite{10411033}, and high-dimensional deformation \cite{zhou2024imitating}. These characteristics make perception and planning significantly more difficult compared to rigid-object manipulation. Traditional approaches rely on geometric cues or template matching \cite{doumanoglou2016folding}, while reinforcement learning methods \cite{wang2024rl} attempt to discover folding strategies through trial and error, often requiring extensive interaction. Learning-based perception techniques \cite{tian2025diffusion} have been proposed for cloth state estimation or mesh reconstruction, but remain disconnected from semantic task goals. Language-conditioned methods \cite{chen2025metafold} enable simple instruction following, but typically depend on fixed templates or supervised mapping, limiting their ability to interpret abstract commands. Unlike rigid-object manipulation, cloth folding demands semantically meaningful, sequential operations grounded in dynamic visual states. In this work, we address these challenges by combining an LLM-based planner for decomposing high-level instructions with a VLM-based module for sub-task grounding to achieve generalizable language-driven cloth manipulation.

\section{Methods}
\subsection{Problem Formulation}
At time step \( t \), given a high-level user command \( S \) and an RGB-D observation \( o_t \in \mathbb{R}^{C \times H \times W} \), the goal is to manipulate a cloth into the folded configuration described by the command. The process involves four main modules: an ASR module \( f_{\text{asr}} \), a Task Planner \( \mathcal{M}_T \), a Visual Module \( \mathcal{M}_V \), and an Action Executor \( \mathcal{M}_A \). First, the ASR module transcribes the spoken input into natural language as \( L_q = f_{\text{asr}}(S) \). Next, the Task Planner \( \mathcal{M}_T \) parses \( L_q \) into a sequence of fine-grained sub-tasks \( L = (\ell_1, \ell_2, \dots, \ell_n) \), where each \( \ell_t \) describes an atomic manipulation step.  At each step \( t \), the Visual Module \( \mathcal{M}_V \) takes the current observation \( o_t \) and sub-task instruction \( \ell_t \) as input, and predicts a corresponding low-level action \( a_t = \mathcal{M}_V(o_t, \ell_t) \). The Action Executor \( \mathcal{M}_A \) then executes \( a_t \), resulting in a new observation \( o_{t+1} \). This perception-action loop continues until all sub-tasks are completed and the target folded configuration is achieved.  The central challenge lies in grounding sequential natural language instructions into accurate and reliable robotic actions, despite the highly deformable nature of cloth. Algorithm 1 outlines the core procedure of our proposed framework.

\begin{figure}[t]
\centering
\includegraphics[width=1\linewidth]{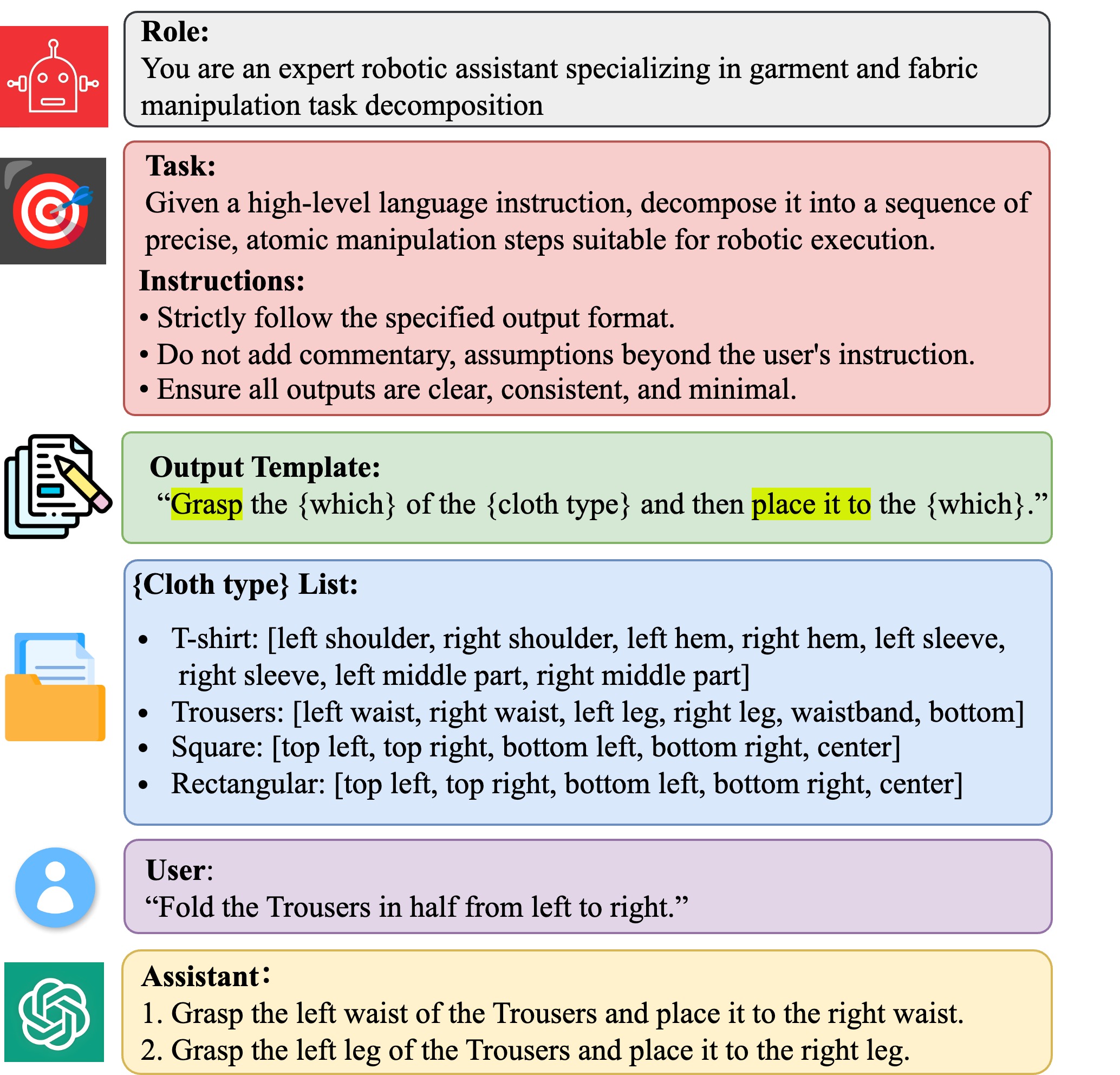}
\caption{\textbf{Task Decomposition.} Example of task decomposition from a high-level instruction using LLM-based planner.}
\end{figure}

\subsection{Task Planning Module}

The Task Planning Module \( \mathcal{M}_T \) addresses the challenge of executing long-horizon tasks that require contextual language understanding and precise action decomposition. Given a language instruction and a textual scene description, an LLM (GPT-4o) decomposes the user command \( L_q \) into a sequence of sub-tasks \( L = (\ell_1, \ell_2, \dots, \ell_n) \). The decomposition process is facilitated through carefully crafted prompts incorporating in-context instruction learning (ICIL) and chain-of-thought (CoT) reasoning, ensuring that each sub-task is standardized and aligned with a set of robot-executable primitives such as pick-and-place operations.

Figure 2 illustrates a representative example of this process for the instruction ``\textit{Fold the Trousers in half from left to right''}. The planner outputs two atomic actions: ``\textit{Grasp the left waist of the Trousers and place it to the right waist''} and ``\textit{Grasp the left leg of the Trousers and place it to the right leg''}. 
This structured decomposition strategy yields interpretable and scalable solutions for long-horizon garment manipulation, inspired by prior work on LLM-based planning \cite{obata2024lip} \cite{zhao2025learning}.

\begin{figure*}[t]
\centering
\includegraphics[width=1\linewidth]{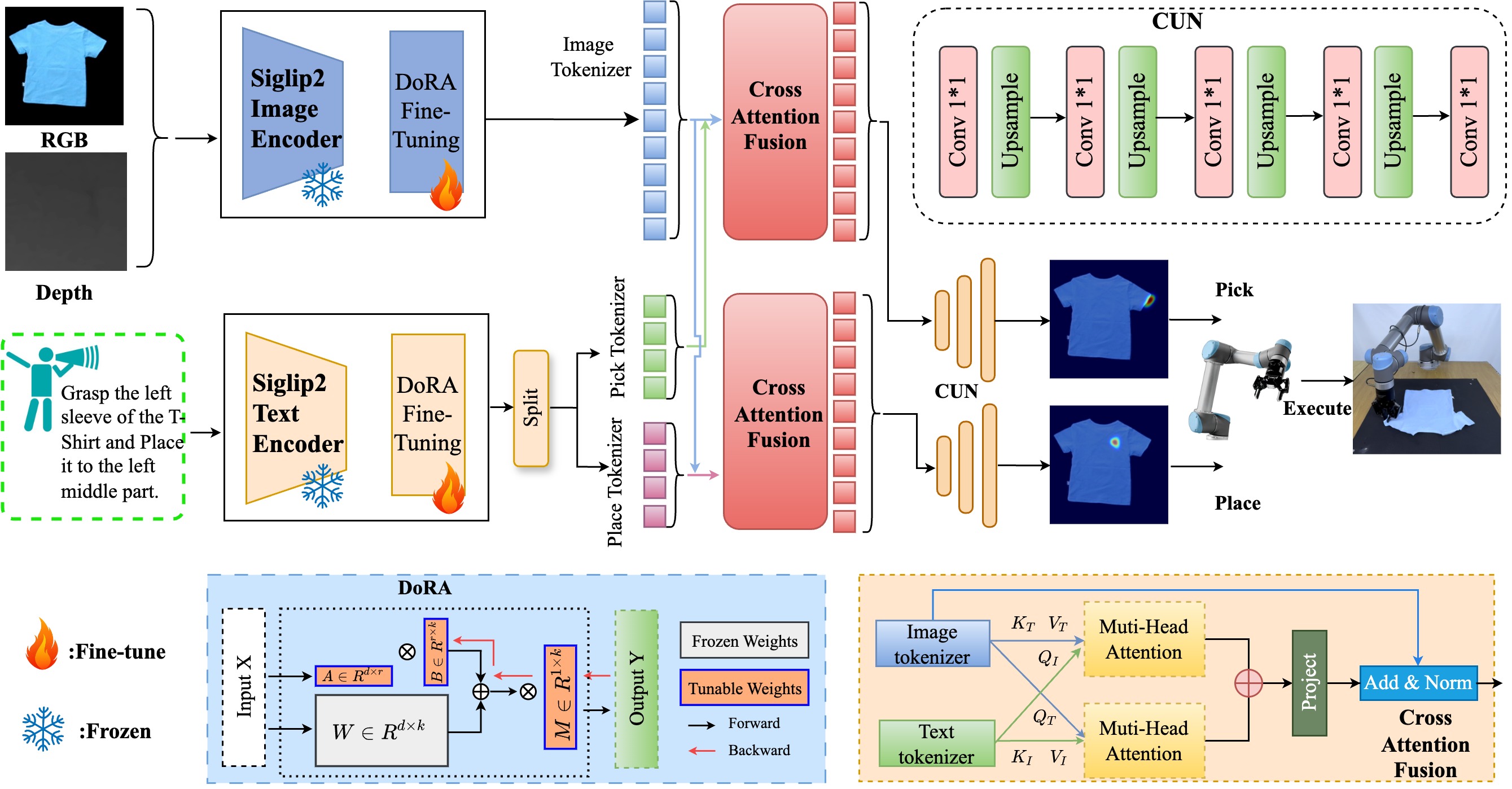}
\caption{\textbf{Architecture of the Visual Perception Module.} The Visual Perception module uses a frozen SigLIP2 model to extract tokens from an RGB-D image and a natural language instruction. The instruction is split at the conjunction ``and'' into pick and place segments. Each segment is fused with visual tokens via bidirectional cross-attention, where textual and visual features are jointly aligned. To adapt the frozen SigLIP2 model to cloth manipulation, DoRA is employed for efficient fine-tuning. The fused features are decoded through convolutional and upsampling layers to predict the corresponding pick and place positions for precise cloth manipulation.}
\end{figure*}

\subsection{Visual Perception Module}
Given an RGB-D scene image $o$ and an object-specific instruction $\ell_t$ from \( \mathcal{M}_T \), \( \mathcal{M}_V \) leverages Grounding DINO \cite{liu2024grounding} to detect garment-related bounding boxes, which are then used to prompt the Segment Anything Model (SAM) \cite{kirillov2023segment} for precise segmentation. The resulting masks isolate garment-relevant regions from the original image, followed by cropping within a predefined workspace. The masked visual input, combined with $\ell_t$, is processed by a vision-language encoder to infer the corresponding manipulation point (see Figure 3). The module comprises three components: multi-modal feature encoding, cross-modal fusion, and action decoding.

\textbf{Multi-Modal Feature Encoding. } We adopt the frozen SigLIP2 model, a contrastive vision-language encoder pretrained on large-scale image-caption pairs. It yields strong visual features and precise language-vision alignment, outperforming prior backbones \cite{li2023blip, radford2021learning}. To tailor it to cloth manipulation while preserving pretrained knowledge, we apply DoRA to the query and value matrices of the attention layers, enabling efficient low-rank updates without altering the full weight parameters.

Specifically, given a sub-task instruction and an RGB-D image $o$, SigLIP2 encodes them as:
\begin{align}
\{ f_\ell, f_o \} = \mathcal{M}_{\text{SigLIP2}}(L_a^n, o)
\end{align}
where $f_\ell \in \mathbb{R}^{B \times T \times D}$ and $f_o \in \mathbb{R}^{B \times P \times D}$ denote token-level language features and patch-level visual features, respectively. Here, $B$ is the batch size, $T$ and $P$ are the number of tokens and image patches, and $D$ is the embedding dimension.

To encourage each decoder to focus on its respective sub-task, we segment the token-level language features $f_\ell$ based on the coordinating conjunction ``\textit{and}''. For instance, given the instruction ``\textit{Grasp the left leg and place it over the right leg}'', we split $f_\ell$ into two segments, the tokens before ``and'' as $f_\ell^1$, and those after as $f_\ell^2$. These segments are then used to provide language grounding for the pick and place decoding branches, respectively. To encode modality-specific priors and enhance representational alignment, we prepend each input with a learnable token: $t_\ell^1, t_\ell^2 \in \mathbb{R}^{1 \times D}$ for the pick and place language branches, and a shared visual token $t_o \in \mathbb{R}^{1 \times 1 \times D}$. These tokens are concatenated with their corresponding modality features to construct the final input representations:
\begin{align}
\mathcal{F}_\ell^1 = [t_\ell^1 \, \Vert \, f_\ell^1], \quad \mathcal{F}_\ell^2 = [t_\ell^2 \, \Vert \, f_\ell^2], \quad \mathcal{F}_o = [t_o \, \Vert \, f_o]
\end{align}
where $\mathcal{F}_\ell^1 \in \mathbb{R}^{B \times (T_1+1) \times D}$ and $\mathcal{F}_\ell^2 \in \mathbb{R}^{B \times (T_2+1) \times D}$ denote the language inputs for the pick and place decoders, respectively; $\mathcal{F}_o \in \mathbb{R}^{B \times (P+1) \times D}$ is the shared visual input; and $T_1$ and $T_2$ represent the number of tokens in each language segment.

\textbf{Cross-modal fusion. } Based on the visual features $\mathcal{F}_o$ and the segmented language features $\mathcal{F}_\ell^1$ and $\mathcal{F}_\ell^2$, we propose a bidirectional cross-attention mechanism to fuse information across modalities. For the pick branch, the image-to-language attention is computed by projecting the inputs with learnable parameters $\mathcal{W}_o^Q$, $\mathcal{W}_o^K$, and $\mathcal{W}_o^V$:
\begin{align}
Q_o = \mathcal{F}_o \mathcal{W}_o^Q, \quad K_o = \mathcal{F}_\ell^1 \mathcal{W}_o^K, \quad V_o = \mathcal{F}_\ell^1 \mathcal{W}_o^V,
\end{align}
\begin{align}
S^{\text{vis}} = \text{softmax}\left( \frac{Q_o K_o^\top}{\sqrt{d_h}} \right) V_o
\end{align}
Similarly, language-to-image attention is computed with a separate set of learnable weights $\mathcal{W}_\ell^Q$, $\mathcal{W}_\ell^K$, and $\mathcal{W}_\ell^V$:
\begin{align}
Q_\ell = \mathcal{F}_\ell^1 \mathcal{W}_\ell^Q, \quad K_\ell = \mathcal{F}_o \mathcal{W}_\ell^K, \quad V_\ell = \mathcal{F}_o \mathcal{W}_\ell^V,
\end{align}
\begin{align}
S^{\text{text}} = \text{softmax}\left( \frac{Q_\ell K_\ell^\top}{\sqrt{d_h}} \right) V_\ell
\end{align}
The two attention outputs are concatenated and projected back to the original dimensionality:
\begin{align}
\mathcal{F}_{\text{fused}} = [S^{\text{vis}} \, \Vert \, S^{\text{text}}] \mathcal{W}^P
\end{align}
Finally, we add a residual connection from the original image features and apply layer normalization:
\begin{align}
\hat{\mathcal{F}}_{\text{pick}} = \text{LayerNorm}(\mathcal{F}_{\text{fused}} + \mathcal{F}_o)
\end{align}
Similarly, the fused feature for the place branch is denoted as $\hat{\mathcal{F}}_{\text{place}}$.

\textbf{Action Decoding. }We design two identical C‌onvolutional ‌U‌psampling ‌N‌etwork (CUN) decoders—one for picking and one for placing—each consisting of stacked $1 \times 1$ convolutional layers interleaved with bilinear upsampling. Given the fused features $\hat{\mathcal{F}}_{\text{pick}}$ and $\hat{\mathcal{F}}_{\text{place}}$, the decoders apply a sigmoid activation to produce the spatial probability maps $\mathcal{Q}_{\text{pick}}$ and $\mathcal{Q}_{\text{place}}$, respectively.
\begin{align}
\mathcal{Q}_{\text{pick}} = \sigma\left(\phi_{\text{pick}}(\hat{\mathcal{F}}_{\text{pick}})\right) \in \mathbb{R}^{H \times W},
\end{align}
\begin{align}
\mathcal{Q}_{\text{place}} = \sigma\left(\phi_{\text{place}}(\hat{\mathcal{F}}_{\text{place}})\right) \in \mathbb{R}^{H \times W}
\end{align}
where $\phi_{\text{pick}}$ and $\phi_{\text{place}}$ denote the CUN decoders and $\sigma(\cdot)$ is the sigmoid function.

The final actions are chosen as the pixel locations with the highest predicted probabilities:
\begin{align}
a_{\text{pick}} = \arg\max_a \mathcal{Q}_{\text{pick}}(a),
\end{align}
\begin{align}
a_{\text{place}} = \arg\max_a \mathcal{Q}_{\text{place}}(a)
\end{align}

\subsection{ Action Translation and Execution}

Given the predicted pick and place locations, pixel coordinates are first projected into 3D world coordinates using depth data and camera intrinsics, and then transformed into the robot base frame via extrinsic calibration. The Action Execution Module then sequentially performs three primitives—grasp (closing the gripper at the pick point to firmly grasp the target region), move-to-position (preserving grasp stability while transporting the cloth segment from the pick point to the place point), and place (releasing the cloth at the place point with precise positioning to achieve the desired configuration).

\subsection{Implementation Details}

We collect expert demonstrations by executing sub-instructions in the SoftGym simulator, yielding 15,750 demonstrations (15,000 for training and 750 for testing).
The dataset is denoted as $\mathcal{D} = \{\zeta_1, \zeta_2, \ldots, \zeta_n\}$, where each trajectory $\zeta_i$ is a sequence of tuples:
\begin{equation}
\zeta_i = \left\{(o_1, l_1, a_1), (o_2, l_2, a_2), \ldots\right\}
\end{equation}
where $o_t$, $l_t$, and $a_t$ denote the observation, the language instruction, and the action at time $t$, respectively. Each $a_t$ is converted into spatial supervision signals represented by ground-truth heatmaps $\mathcal{Q}_{\text{pick}}^{gt}$ and $\mathcal{Q}_{\text{place}}^{gt}$.

The VLM is trained using a binary cross-entropy loss(BCE) between $\mathcal{Q}{\text{pick}}$, $\mathcal{Q}{\text{place}}$ and their ground-truth heatmaps:
\begin{equation}
\mathcal{L}_{\text{total}} = \mathcal{L}_{\text{BCE}}(\mathcal{Q}_{\text{pick}}, \mathcal{Q}_{\text{pick}}^{gt}) + \mathcal{L}_{\text{BCE}}(\mathcal{Q}_{\text{place}}, \mathcal{Q}_{\text{place}}^{gt})
\end{equation}
Each BCE loss is computed as:
\begin{equation}
\begin{split}
\mathcal{L}_{\text{BCE}}(\mathcal{Q}, \mathcal{Q}^{gt}) = -\sum_{i,j} \big[ 
    & \mathcal{Q}^{gt}_{i,j} \log(\mathcal{Q}_{i,j}) \\
    & + (1 - \mathcal{Q}^{gt}_{i,j}) \log(1 - \mathcal{Q}_{i,j}) 
\big]
\end{split}
\end{equation}

The VLM is trained for 100 epochs using the Adam optimizer with a batch size of 16 and a learning rate of 1e-4 on a workstation running Ubuntu 18.04 LTS, equipped with an NVIDIA RTX 4090 GPU, Intel i9-13900K CPU (5.80 GHz), 64 GB RAM, CUDA 12.0, and Docker 19.03.

\section{Simulation Experiments}
\subsection{Simulation Setup}
Our simulation environment is based on the OpenAI Gym API and PyFleX bindings to NVIDIA FleX, integrated via SoftGym \cite{lin2021softgym}. It supports loading arbitrary cloth meshes, such as T-Shirts and Trousers, through a Python interface. The robot gripper is modeled as a spherical picker that moves freely in 3D space and attaches to the nearest cloth particle upon activation. Visual observations are rendered using  OpenGL, producing RGB-D images at a resolution of 224$\times$224.

\subsection{Tasks and Metrics}

We use the following three metrics to assess performance:

\textbf{1) Mean Particle Distance (MPD, m):} The average Euclidean distance between particles in the final cloth state and those in the ground-truth configuration.

\textbf{2) Mean Intersection over Union (MIoU):} The average IoU between the predicted cloth mask and the ground-truth mask across all test cases.

\textbf{3) Success Rate (SR):} A manipulation is considered successful if the MPD is below 0.0125 meters.

These metrics are evaluated on five cloth manipulation tasks: Double Straight Fold (TSF), folding the cloth into a rectangle; Double Triangle Fold (DTF), folding the cloth to form a triangle; Four Corners Inward Fold (FCIF), folding all corners to the center; T-Shirt Fold (TSF), folding the sleeves to the center and then the bottom to the shoulder; and Trousers Fold (TF), folding the legs in half lengthwise and then folding again to double. In addition, evaluations are conducted under three instruction conditions: seen instructions (SI), unseen instructions (UI), and unseen tasks (UT).

\begin{table*}[t]
\centering
\setlength{\tabcolsep}{2.5pt} % 列间距
\renewcommand{\arraystretch}{1.5} % 行距
\caption{Simulation experiment results. Performance (SR (\%), MPD (m), MIoU (\%)) on five cloth-folding tasks under different instruction conditions. Best results are in bold.}
\label{tab:sim-results}
\begin{tabular*}{\textwidth}{@{\extracolsep{\fill}}lccccccccccccccccccc@{}}
\toprule
\multirow{2}{*}{Task} & \multirow{2}{*}{Method} & \multicolumn{3}{c}{DSF} & \multicolumn{3}{c}{DTF} & \multicolumn{3}{c}{FCIF} & \multicolumn{3}{c}{TF} & \multicolumn{3}{c}{TSF} & \multirow{2}{*}{Avg} \\
\cmidrule(lr){3-5} \cmidrule(lr){6-8} \cmidrule(lr){9-11} \cmidrule(lr){12-14} \cmidrule(lr){15-17}
& & SR & MPD & MIoU & SR & MPD & MIoU & SR & MPD & MIoU & SR & MPD & MIoU & SR & MPD & MIoU & \\

\hline
\multirow{3}{*}{SI}
& Deng et al. & 81.3 & 0.0077 & 91.60 & 80.0 & 0.0076 & 89.08 & 80.0 & 0.0143 & 87.87 & 76.0 & 0.0100 & 90.15 & 85.0 & 0.0083 & 92.95 & 80.46 \\
& Barbany et al. & 85.0 & 0.0079 & 91.38 & 93.0 & 0.0049 & 92.11 & \textbf{100.0} & 0.0048 & 94.44 & 77.33 & 0.0094 & 90.83 & \textbf{92.0} & 0.0075 & 93.78 & 89.47 \\
& \textbf{Ours} & \textbf{89.0} & \textbf{0.0067} & \textbf{92.58} & \textbf{94.0} & \textbf{0.0050} & \textbf{92.25} & \textbf{100.0} & \textbf{0.0045} & \textbf{94.68} & \textbf{84.0} & \textbf{0.0088} & \textbf{91.57} & 91.5 & \textbf{0.0074} & \textbf{93.80} & \textbf{91.70} \\

\hline
\multirow{3}{*}{UI}
& Deng et al. & 79.0 & 0.0093 & 90.75 & 70.0 & 0.0151 & 84.74 & 98.0 & 0.0046 & 95.06 & 74.0 & 0.0104 & 90.37 & 88.0 & 0.0079 & 93.32 & 81.80 \\
& Barbany et al. & 84.0 & 0.0084 & 91.38 & 91.5 & 0.0053 & 91.88 & \textbf{100.0} & \textbf{0.0031} & \textbf{96.20} & 72.66 & 0.0104 & 90.46 & 90.5 & 0.0071 & 94.04 & 87.73 \\
& \textbf{Ours} & \textbf{88.0} & \textbf{0.0075} & \textbf{91.66} & \textbf{94.0} & \textbf{0.0048} & \textbf{92.49} & 97.0 & 0.0062 & 92.96 & \textbf{75.5} & \textbf{0.0095} & \textbf{91.04} & \textbf{93.5} & \textbf{0.0070} & \textbf{94.21} & \textbf{89.60} \\

\hline
\multirow{3}{*}{UT}
& Deng et al. & 66.0 & 0.0375 & 77.44 & 0 & 0.0100 & 55.18 & 98.0 & 0.0050 & 94.48 & 0 & 0.0758 & 48.98 & 50.0 & 0.0343 & 94.60 & 42.80 \\
& Barbany et al. & 68.6 & 0.0306 & 77.90 & 0 & 0.0870 & 50.45 & \textbf{100.0} & \textbf{0.0038} & \textbf{95.36} & 3.33 & 0.0965 & 36.91 & 51.0 & 0.0378 & 69.68 & 44.59 \\
& \textbf{Ours} & \textbf{82.6} & \textbf{0.0072} & \textbf{90.20} & \textbf{69.0} & \textbf{0.0108} & \textbf{85.27} & \textbf{100.0} & 0.0059 & 93.16 & \textbf{70.0} & \textbf{0.0123} & \textbf{88.02} & \textbf{68.0} & \textbf{0.0108} & \textbf{90.65} & \textbf{77.92} \\
\bottomrule
\end{tabular*}
\end{table*}

\begin{figure*}[t]
\centering
\includegraphics[width=1\linewidth]{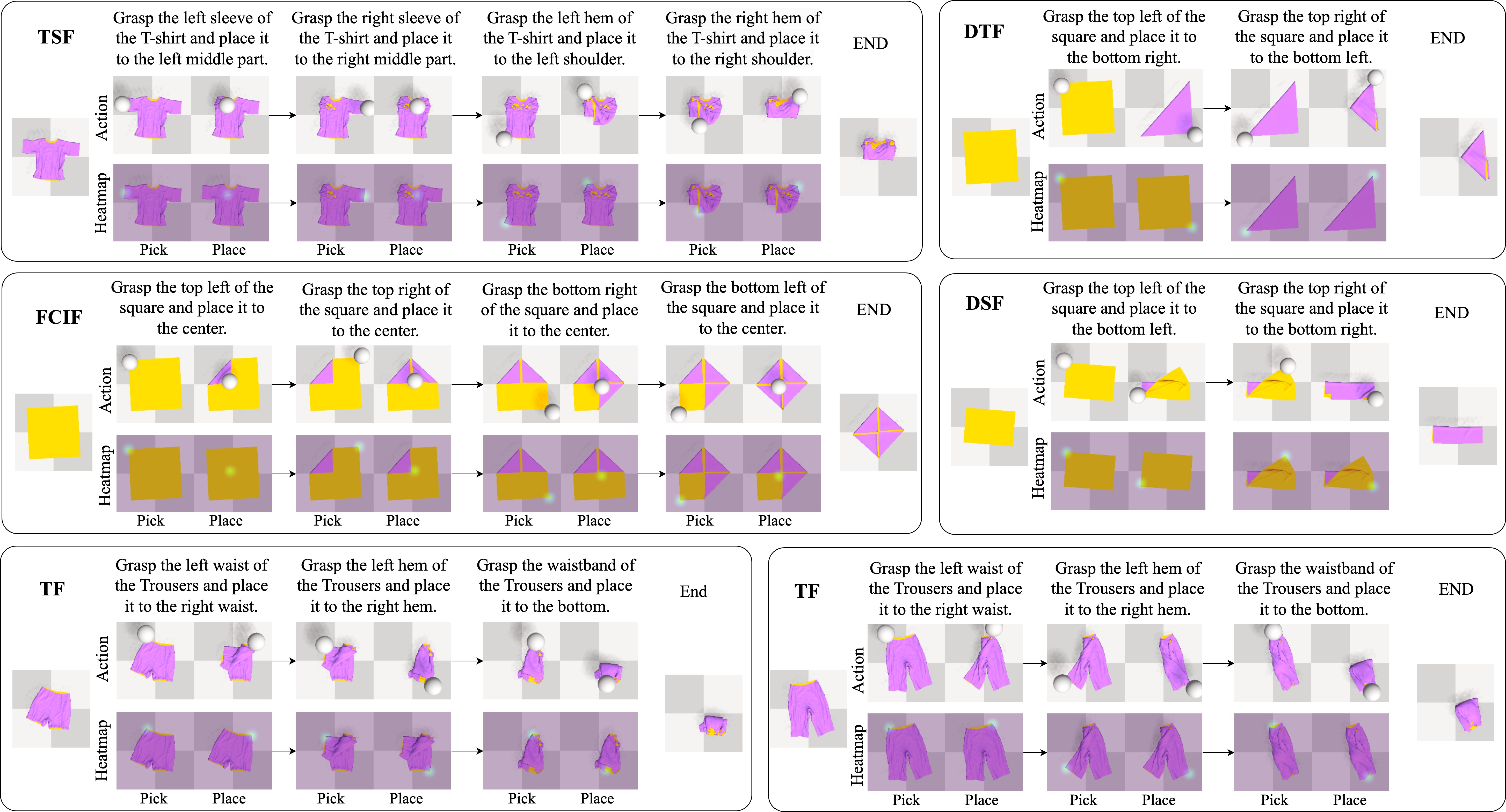}
\caption{\textbf{Qualitative Simulation Results on Five Folding Tasks with See Instructions.}  Each task is represented in two rows: the top row shows the top-view action sequence captured by the overhead camera, and the bottom row displays the network-predicted pick and place heatmaps.}
\end{figure*}

\subsection{Baseline Methods}
To evaluate our method’s manipulation capabilities, we compare it against two baselines: Deng et al. \cite{deng2024generalizable} and Barbany et al. \cite{barbany2025bifold}. Both are language-guided folding methods specifically developed for deformable object manipulation. Barbany et al. represent the current SOTA method, and we adapt it to a single-arm variant for consistency with our setting. For a fair comparison, we reproduce both baselines and train them on same dataset as ours, following their respective fine-tuning and model updating strategies.

\begin{table*}[t]
\centering
\small
\caption{
Ablation study of architecture modules and backbones (\%). Best results are in bold.
}
\renewcommand{\arraystretch}{1.3}
\resizebox{\textwidth}{!}{%
\begin{tabular}{l l ccc ccc ccc ccc ccc c}
\toprule
\multirow{2}{*}{\textbf{Backbone}} & \multirow{2}{*}{\textbf{Fusion}}  
& \multicolumn{3}{c}{\textbf{DSF}} 
& \multicolumn{3}{c}{\textbf{DTF}} 
& \multicolumn{3}{c}{\textbf{FCIF}} 
& \multicolumn{3}{c}{\textbf{TF}} 
& \multicolumn{3}{c}{\textbf{TSF}} 
    &\multirow{2}{*}{ \textbf{Avg}}\\
\cmidrule(lr){3-5}
\cmidrule(lr){6-8}
\cmidrule(lr){9-11}
\cmidrule(lr){12-14}
\cmidrule(lr){15-17}
& & SI & UI & UT & SI & UI & UT & SI & UI & UT & SI & UI & UT & SI & UI & UT & \\
\midrule
CLIP & Cross-Attn    
& 85.0 & 83.0 & 79.3& 86.5 & 84.0 & 61.4& 96.5 & 95.0 & 91.0 
& 68.0 & 73.0 & 65.0 
& 79.5 & 82.0 & 66.6& 79.72\\
BLIP2   & Cross-Attn    
& 87.5 & 85.0 & 80.0& 90.0& 93.5& 66.6& \textbf{100.0} & 99.9 & 94.0 & 75.6& 77.3 & 69.7 & 82.6& 79.0& 69.0 & 83.31\\
SigLIP2 & Transformer   
& 84.0 & \textbf{88.0} & \textbf{82.6}& 93.0 & 92.0 & 63.0 
& \textbf{100.0} & \textbf{100.0} & 97.5 
& 74.0 & \textbf{81.3} & \textbf{72.0 }
& 87.5 & 82.5 &\textbf{ 68.5 }
& 84.39\\
\textbf{Ours(SigLIP2)} & \textbf{Cross-Attn} 
& \textbf{89.0} & \textbf{88.0} & \textbf{82.6} 
& \textbf{94.0} & \textbf{94.0} & \textbf{69.0} 
& \textbf{100.0} & 97.0 & \textbf{100.0} 
& \textbf{84.0} & 75.5 & 70.0 
& \textbf{91.5} & \textbf{93.5} & 68.0 
& \textbf{86.40} \\
\bottomrule
\end{tabular}%
}
\end{table*}\begin{table*}[t]
\centering
\small
\caption{Ablation study of parameter-efficient fine-tuning modules (\%). Best results are in bold. }
\renewcommand{\arraystretch}{1.3}
\resizebox{\textwidth}{!}{%
\begin{tabular}{ccc ccc ccc ccc ccc ccc c}
\toprule
\multirow{2}{*}{\textbf{LoRA}} & \multirow{2}{*}{\textbf{IA$^3$}} & \multirow{2}{*}{\textbf{DoRA}} 
& \multicolumn{3}{c}{\textbf{DSF}} 
& \multicolumn{3}{c}{\textbf{DTF}} 
& \multicolumn{3}{c}{\textbf{FCIF}} 
& \multicolumn{3}{c}{\textbf{TF}} 
& \multicolumn{3}{c}{\textbf{TSF}} 
& \multirow{2}{*}{\textbf{Avg}} \\
\cmidrule(lr){4-6}
\cmidrule(lr){7-9}
\cmidrule(lr){10-12}
\cmidrule(lr){13-15}
\cmidrule(lr){16-18}
& & & SI & UI & UT & SI & UI & UT & SI & UI & UT & SI & UI & UT & SI & UI & UT & \\
\midrule
\ding{55} & \ding{55} & \ding{55} & 85.0& 69.0 & 74.0 & 87.0 & 79.0 & 65.0 & 85.0 & 96.0 & 91.5 & 64.6 & 69.3 & 61.3 & 75.5 & 74.0 & 54.5 &75.38\\
\ding{51} & \ding{55} & \ding{55} & 87.0 & \textbf{88.0} & 83.0 & 91.0 & 93.0 & 63.3 & 99.5 & \textbf{100.0} & 99.5 & 79.3 & \textbf{82.6} & \textbf{71.3} &87.0 & 83.5 & 66.6 & 84.97 \\
\ding{55} & \ding{51} & \ding{55} & 84.0 & \textbf{88.0} & 79.3 & 88.0 & 89.0 & 64.0 & 99.0 & \textbf{100.0} & \textbf{100.0} & 77.3 & 75.3 & 68.6 & 86.5 & 85.0 & 57.9 &82.79\\
\textbf{\ding{55}} & \textbf{\ding{55}} & \textbf{\ding{51}} 
& \textbf{89.0} & \textbf{88.0} & \textbf{82.6} 
& \textbf{94.0} & \textbf{94.0} & \textbf{69.0} 
& \textbf{100.0} & 97.0 & \textbf{100.0} 
& \textbf{84.0} & 75.5 & 70.0
& \textbf{91.5} & \textbf{93.5} & \textbf{68.0} 
& \textbf{86.40} \\
\bottomrule
\end{tabular}%
}
\end{table*}\begin{figure}[t]
\centering
\includegraphics[width=1\linewidth]{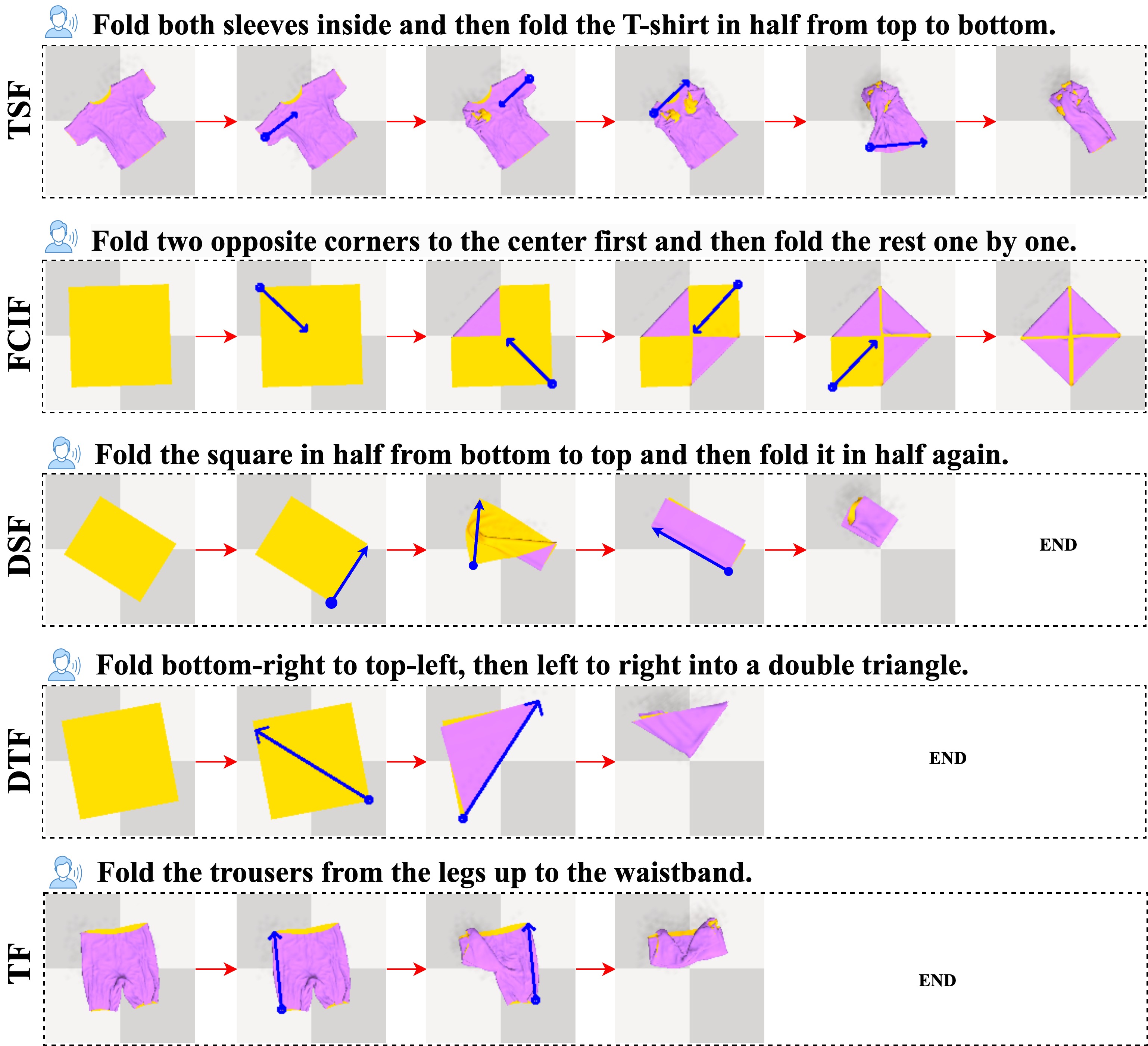}
\caption{\textbf{Qualitative results on unseen tasks in simulation.} Each row shows a multi-step folding sequence for a different cloth type, performed under the \textit{unseen task} setting.}
\end{figure}

\subsection{Results and Analysis}
Table I shows the quantitative results of our method across five cloth manipulation tasks under three instruction conditions.  Our approach achieves the highest performance across all three metrics—MPD, MIoU, and SR—consistently surpassing existing baselines. Under seen instructions, our method achieves the highest average SR (91.7\%), outperforming Barbany et al. (89.47\%) and Deng et al. (80.46\%). Under the unseen instructions condition,  which evaluates the model's ability to generalize to novel language commands, our approach continues to perform well, reaching 89.6\%, exceeding Deng et al. (81.8\%) and slightly surpassing Barbany et al. (87.73\%). This demonstrates the model’s robustness to linguistic variation. In the unseen tasks setting—the most challenging of the three—our method achieves a 77.92\% SR, substantially outperforming Barbany et al.  (44.59\%) and Deng et al. (42.8\%). This demonstrates the strong generalization ability of our method, particularly in complex manipulations such as TF and TSF, where baseline methods struggle or fail to complete the task. We further provide qualitative results in Figure 4 and Figure 5, which visualizes the multi-step task execution process. With language instructions as input, our model predicts where to manipulate the cloth and completes the folds. It performs well not only on seen instructions, but also generalizes to new instructions and unseen tasks.

\begin{table}[t]
\centering
\renewcommand{\arraystretch}{1.1} % 行距
\caption{Ablation study results of LLMs (\%).}
\small
\setlength{\tabcolsep}{4.5pt} % 列间距
\begin{tabular}{lccccc}
\toprule
Condition & DSF & DTF & FCIF & TF & TSF \\
\midrule
Doubao-1.5   & 80.0 & 60.0 & 20.0 & 40.0 & 20.0 \\
Grok-3       & 100.0 & 80.0 & 60.0 & 100.0 & 40.0 \\
DeepSeek-V3  & 100.0 & 100.0 & 100.0 & 80.0 & 80.0 \\
GPT-4o       & 100.0 & 100.0 & 100.0 & 100.0 & 100.0 \\
\bottomrule
\end{tabular}
\end{table}

\subsection{Ablation Study}
We conduct ablations to evaluate the effectiveness of the net architecture and parameter-efficient fine-tuning.

\textbf{Backbone and Fusion Strategy.} As shown in Table II, we compare three vision-language backbones (SigLIP2 \cite{tschannen2025siglip}, BLIP2 \cite{li2023blip}, and CLIP \cite{radford2021learning}) and two fusion strategies: cross-attention and transformer-based fusion. Our full model Talk2Fold(SigLIP2) achieves the highest average SR of 86.40\% across all 15 subtasks. Replacing the cross-attention module with a transformer-based fusion leads to a performance drop of 2.01\%, highlighting the effectiveness of cross-attention in extracting language-guided visual features. Compared to CLIP and BLIP2, SigLIP2 consistently outperforms both, demonstrating superior generalization capabilities.

\begin{table*}[t]
\centering
\setlength{\tabcolsep}{4pt} % 列间距
\renewcommand{\arraystretch}{1.35} % 行距
\caption{Real-world experiment results. Performance (SR (\%), MIoU (\%)) on five cloth-folding tasks under different instruction conditions. Best results are in bold.}
\label{tab:real-results}
\begin{tabular*}{\textwidth}{@{\extracolsep{\fill}}lcccccccccccc@{}}
\toprule
\multirow{2}{*}{Task} & \multirow{2}{*}{Method} & \multicolumn{2}{c}{DSF} & \multicolumn{2}{c}{DTF} & \multicolumn{2}{c}{FCIF} & \multicolumn{2}{c}{TF} & \multicolumn{2}{c}{TSF} & \multirow{2}{*}{Avg} \\
\cmidrule(lr){3-4} \cmidrule(lr){5-6} \cmidrule(lr){7-8} \cmidrule(lr){9-10} \cmidrule(lr){11-12}
& & SR & MIoU & SR & MIoU & SR & MIoU & SR & MIoU & SR & MIoU & \\
\midrule
\multirow{2}{*}{SI}
& Barbany et al. & 80.0 & 85.31 & 80.0 & 87.78 & \textbf{100.0} & \textbf{88.72} & \textbf{80.0} & \textbf{84.18} & \textbf{80.0} & 85.49 & 84.0 \\
& \textbf{Ours} & \textbf{100.0} & \textbf{85.87} & \textbf{100.0} & \textbf{88.57} & 80.0 & 86.30 & \textbf{80.0} & 83.58 & \textbf{80.0} & \textbf{86.67} & \textbf{88.0} \\
\midrule
\multirow{2}{*}{UI}
& Barbany et al. & 60.0 & 82.91 & 80.0 & 82.00 & \textbf{80.0} & \textbf{84.54} & 60.0 & 82.66 & 60.0 & 80.90 & 68.0 \\
& \textbf{Ours} & \textbf{80.0} & \textbf{84.77} & \textbf{100.0} & \textbf{83.35} & \textbf{80.0} & 82.67 & \textbf{80.0} & \textbf{84.08} & \textbf{80.0} & \textbf{83.42} & \textbf{84.0} \\
\midrule
\multirow{2}{*}{UT}
& Barbany et al. & 20.0 & 72.35 & 0.0 & 31.90 & 60.0 & 74.49 & 0.0 & 38.70 & 40.0 & 65.18 & 24.0 \\
& \textbf{Ours} & \textbf{80.0} & \textbf{82.88} & \textbf{60.0} & \textbf{76.26} & \textbf{80.0} & \textbf{79.73} & \textbf{60.0} & \textbf{79.20} & \textbf{60.0} & \textbf{80.65} & \textbf{68.0} \\
\bottomrule
\end{tabular*}
\end{table*}

\textbf{Parameter-Efficient Tuning Modules.} Table III presents the impact of different parameter-efficient tuning strategies: low-rank adaptation (LoRA) \cite{hu2022lora}, IA$^3$ \cite{liu2022few}, and DoRA \cite{liu2024dora}. Among them, using DoRA alone achieves the highest average SR (86.40\%), outperforming LoRA (84.97\%) and IA$^3$ (82.79\%). All tuning methods significantly outperform the no-adaptation baseline (75.38\%), underscoring the importance of tuning the language-conditioned model. Compared to other methods, DoRA provides better performance, making it the most suitable choice in our setting.

\textbf{LLMs.}
    High-level task planning plays a critical role in our cloth manipulation framework. To assess the effectiveness of different LLMs in this component, we benchmark four recent models: Doubao-1.5 \cite{team2025doubao}, Grok-3 \cite{xAIGrok3_2025}, DeepSeek-V3 \cite{liu2024deepseek},  and GPT-4o \cite{hurst2024gpt}. For each model, we conduct 5 trials per task and collect expert annotations to determine the planning SR. As reported in Table IV, GPT-4o achieves the highest average SR, indicating superior reasoning and instruction-following capabilities in this domain. Based on these results, we adopt GPT-4o as the task planner in all subsequent experiments.

\begin{figure}[t]
  \centering
\centering
\includegraphics[width=1\linewidth]{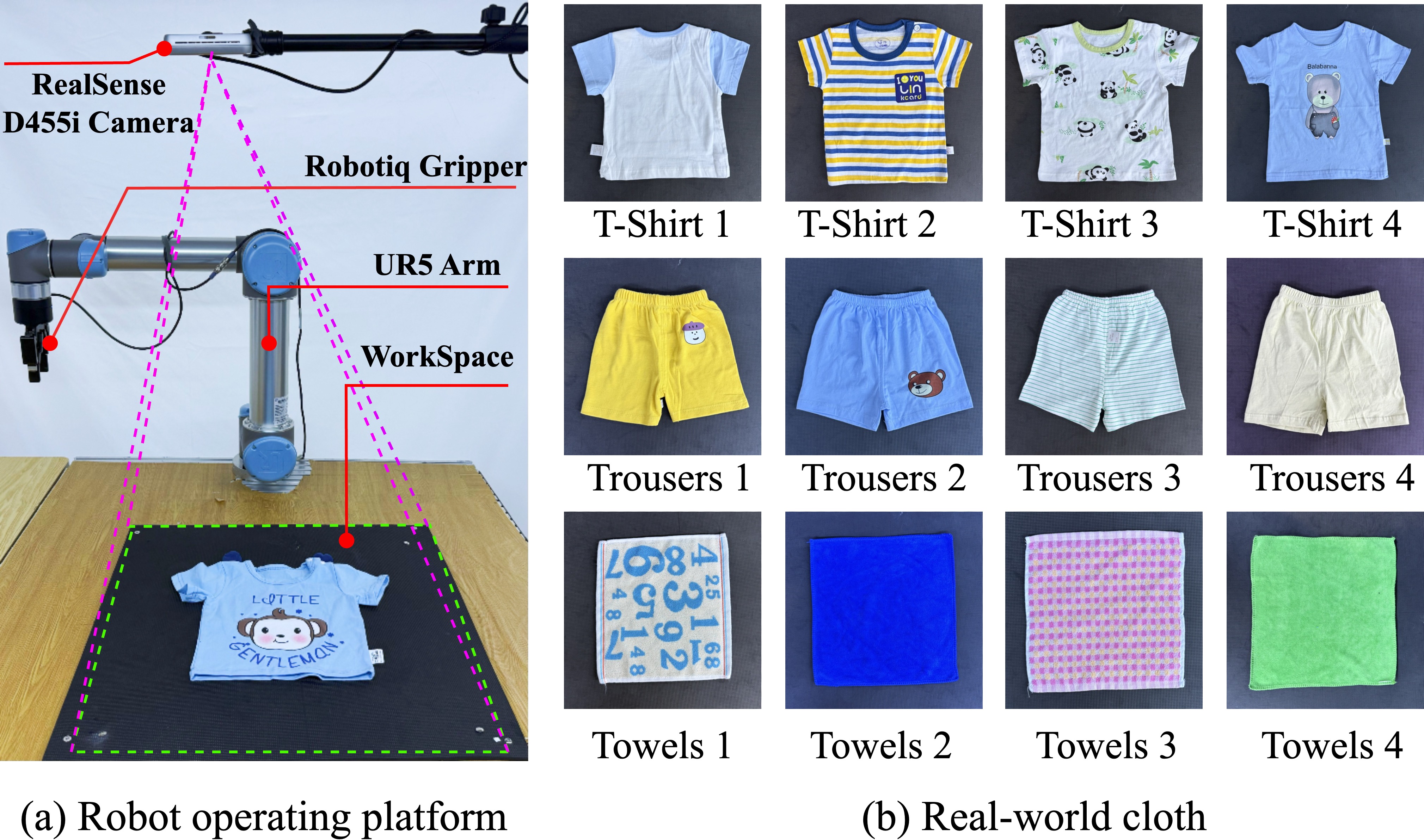}
%\caption{fig2}
\caption{\textbf{Experimental Setup.} Real-world environment and cloth categories used for model testing.}
\label{}
\end{figure}
 
\begin{figure}[t]
\centering
\includegraphics[width=0.5\linewidth]{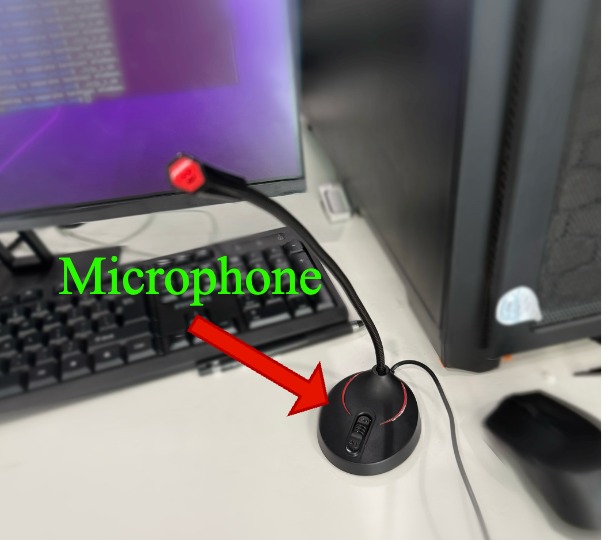}
%\caption{fig2}

\caption{\textbf{Microphone for Speech Input.} The microphone is used to capture user-issued spoken instructions, which are then processed by the ASR module for downstream task planning.}
\label{}
\end{figure}

\section{Real World Experiments}
\subsection{Real World Setup}
Our physical system (see Figure 6) employs Baidu Qianfan’s Automatic Speech Recognition (ASR) interface to convert user speech captured via a microphone (see Figure 7) into text, which is then processed by the LLM to generate a sequence of executable sub-instructions. These are executed by a 6-DoF UR5 robotic arm with a Robotiq 2-Finger Gripper through hand–eye calibration, inverse kinematics, and URScript-based motion primitives. The robot operates on a $1,\mathrm{m} \times 1,\mathrm{m}$ workspace covered with a rigid polypropylene (PP) board, which prevents collisions that could damage the arm. An Intel RealSense D455 RGB-D camera is mounted above the workspace to provide $640 \times 480$ visual observations, which are center-cropped to $224 \times 224$ before being fed into the model. Real-world experiments use 12 clothes with variations in shape, color, and material (see Fig. 6).

\begin{figure}[t]
  \centering
\centering
\includegraphics[width=1\linewidth]{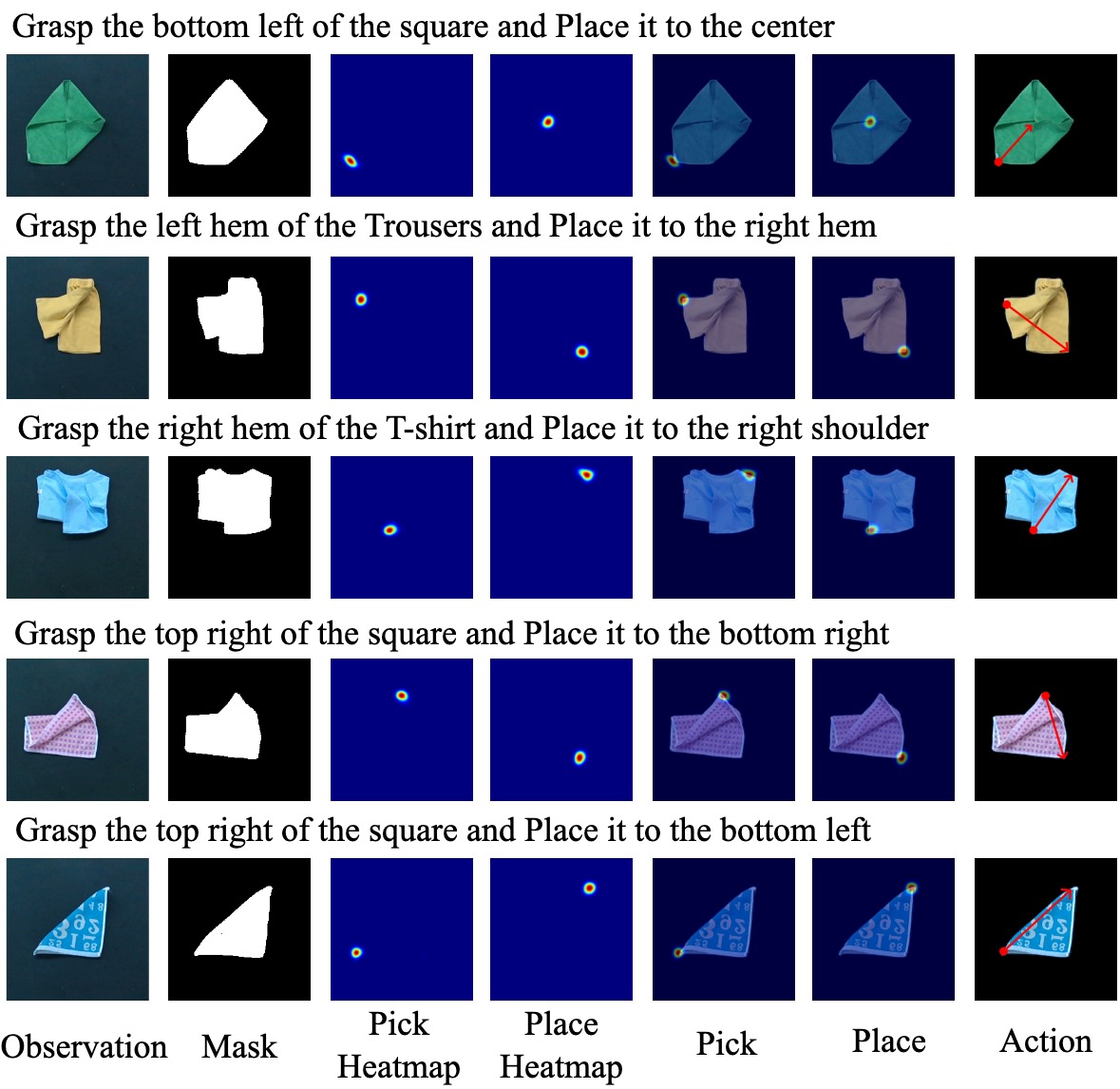}
%\caption{fig2}
\caption{\textbf{Fabric Categories Used for Model Training and Testing.} The dataset includes 18 samples from each of three fabric types: towels, T-shirts, and shorts.}
\label{}
\end{figure}

\begin{figure*}[t]
\centering
\includegraphics[width=1\linewidth]{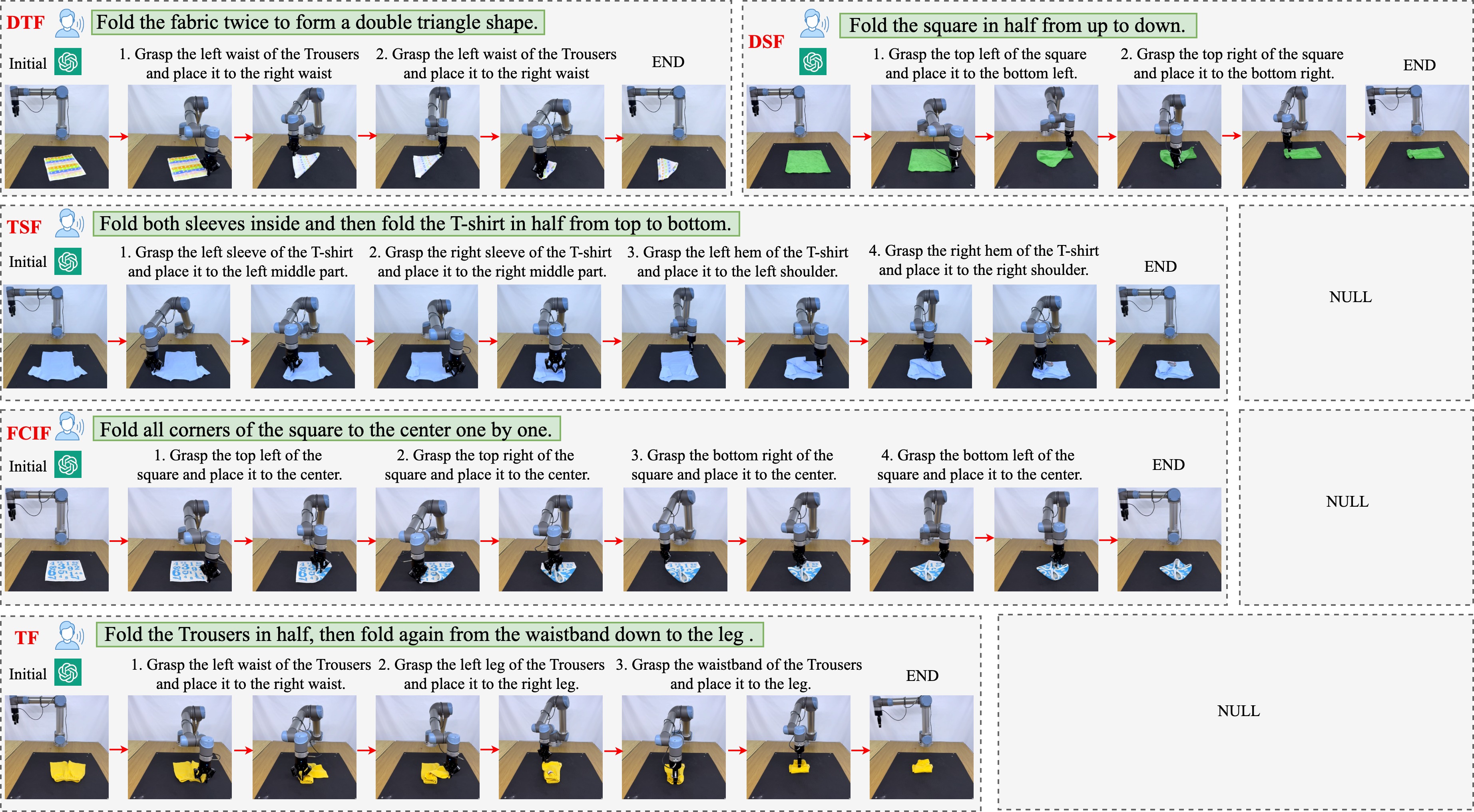}
\caption{\textbf{Qualitative Real-World Results on Five Folding Tasks.} Each task is visualized in three rows: the top row displays the user-provided voice instruction, the middle row shows the subtask breakdown generated by the LLM-based planner, and the bottom row illustrates the robot performing actions according to the decomposed commands.}
\end{figure*}

\subsection{Real-World Experiments}

To evaluate the sim-to-real transfer performance of our approach, we conduct real-world experiments (see Table V) on five language-conditioned cloth manipulation tasks. Each task is executed six times (two trials per instruction variant) on a real robotic platform. The evaluation covers three conditions—Seen Instructions, Unseen Instructions, and Unseen Tasks. We report Success Rate (SR), based on the proportion of successful trials, and Mean IoU (MIoU), which quantifies the accuracy of final cloth alignment. A trial is considered successful if the MIoU between the resulting cloth mask and a human-demonstrated reference exceeds 0.8.

As shown in Table V, our method achieves average success rates of 88\%, 84\%, and 68\% across the three instruction conditions, consistently surpassing the Barbany et al. baseline (84\%, 68\%, and 24\%). These results confirm the robustness of our approach for sim-to-real transfer and its strong generalization to unseen instructions and novel tasks. Qualitative evidence further supports this: Figure 8 illustrates an atomic action prediction at a specific state from a top-down view, demonstrating precise perception across diverse cloth states and fabric types, while Figure 9 showcases the system successfully completing multi-step folding in the real world, guided by user instructions.

\section{Conclusion}

We present a unified framework for language-guided cloth manipulation that integrates speech-based task input, LLM-driven planning, and VLM-based visual grounding. By decomposing high-level instructions into structured sub-tasks and grounding them in RGB-D observations via a bidirectional cross-attention module, the framework executes multi-step cloth folding tasks in both simulated and real-world environments. To adapt to cloth, we introduce DoRA, a parameter-efficient tuning strategy that enhances the perception model’s sensitivity to cloth features. Extensive evaluations demonstrate that the proposed approach outperforms prior SOTA methods under various instruction and task settings, achieving robust performance. These results suggest that integrating LLM-based planning with VLM-based perception provides a viable pathway for language-driven manipulation of deformable objects. Future work will focus on expanding this approach to dual-arm setups and other cloth categories.

{\small
\bibliographystyle{IEEEtran} % 保持 IEEEtran（BibTeX）
\bibliography{Ref}           % 你的 .bib 文件

\begin{thebibliography}{10}
\providecommand{\url}[1]{#1}
\csname url@samestyle\endcsname
\providecommand{\newblock}{\relax}
\providecommand{\bibinfo}[2]{#2}
\providecommand{\BIBentrySTDinterwordspacing}{\spaceskip=0pt\relax}
\providecommand{\BIBentryALTinterwordstretchfactor}{4}
\providecommand{\BIBentryALTinterwordspacing}{\spaceskip=\fontdimen2\font plus
\BIBentryALTinterwordstretchfactor\fontdimen3\font minus \fontdimen4\font\relax}
\providecommand{\BIBforeignlanguage}[2]{{%
\expandafter\ifx\csname l@#1\endcsname\relax
\typeout{** WARNING: IEEEtran.bst: No hyphenation pattern has been}%
\typeout{** loaded for the language `#1'. Using the pattern for}%
\typeout{** the default language instead.}%
\else
\language=\csname l@#1\endcsname
\fi
#2}}
\providecommand{\BIBdecl}{\relax}
\BIBdecl

\bibitem{mon2025embodied}
R.~Mon-Williams, G.~Li, R.~Long, W.~Du, and C.~G. Lucas, ``Embodied large language models enable robots to complete complex tasks in unpredictable environments,'' \emph{Nature Machine Intelligence}, pp. 1--10, 2025.

\bibitem{9097275}
J.~Borràs, G.~Alenyà, and C.~Torras, ``A grasping-centered analysis for cloth manipulation,'' \emph{IEEE Transactions on Robotics}, vol.~36, no.~3, pp. 924--936, 2020.

\bibitem{10611491}
J.~Yang, C.~Deng, J.~Wu, R.~Antonova, L.~Guibas, and J.~Bohg, ``Equivact: Sim(3)-equivariant visuomotor policies beyond rigid object manipulation,'' in \emph{2024 IEEE International Conference on Robotics and Automation (ICRA)}, 2024, pp. 9249--9255.

\bibitem{gu2025tactilealoha}
N.~Gu, K.~Kosuge, and M.~Hayashibe, ``Tactilealoha: Learning bimanual manipulation with tactile sensing,'' \emph{IEEE Robotics and Automation Letters}, 2025.

\bibitem{wu2024unigarmentmanip}
R.~Wu, H.~Lu, Y.~Wang, Y.~Wang, and H.~Dong, ``Unigarmentmanip: A unified framework for category-level garment manipulation via dense visual correspondence,'' in \emph{Proceedings of the IEEE/CVF Conference on Computer Vision and Pattern Recognition}, 2024, pp. 16\,340--16\,350.

\bibitem{diao2025soundmind}
X.~Diao, C.~Zhang, K.~Kong, W.~Wu, C.~Ma, Z.~Ouyang, P.~Qing, S.~Vosoughi, and J.~Gui, ``Soundmind: Rl-incentivized logic reasoning for audio-language models,'' \emph{arXiv preprint arXiv:2506.12935}, 2025.

\bibitem{encoder}
Z.~Li, Z.~Chen, H.~Wen, Z.~Fu, Y.~Hu, and W.~Guan, ``Encoder: Entity mining and modification relation binding for composed image retrieval,'' in \emph{Proceedings of the AAAI Conference on Artificial Intelligence}, vol.~39, no.~5, 2025, pp. 5101--5109.

\bibitem{pmlr-v164-shridhar22a}
M.~Shridhar, L.~Manuelli, and D.~Fox, ``Cliport: What and where pathways for robotic manipulation,'' in \emph{Conference on robot learning}.\hskip 1em plus 0.5em minus 0.4em\relax PMLR, 2022, pp. 894--906.

\bibitem{brohan2023rt}
B.~Zitkovich, T.~Yu, S.~Xu, P.~Xu, T.~Xiao, F.~Xia, J.~Wu, P.~Wohlhart, S.~Welker, A.~Wahid \emph{et~al.}, ``Rt-2: Vision-language-action models transfer web knowledge to robotic control,'' in \emph{Conference on Robot Learning}.\hskip 1em plus 0.5em minus 0.4em\relax PMLR, 2023, pp. 2165--2183.

\bibitem{li2024vision}
X.~Li, M.~Liu, H.~Zhang, C.~Yu, J.~Xu, H.~Wu, C.~Cheang, Y.~Jing, W.~Zhang, H.~Liu \emph{et~al.}, ``Vision-language foundation models as effective robot imitators,'' in \emph{ICLR}, 2024.

\bibitem{deng2024generalizable}
Y.~Deng and D.~Hsu, ``Generalizable clothes manipulation with large language model,'' in \emph{2024 ICRA Workshop on Representing and Manipulating Deformable Objects}, 2024.

\bibitem{barbany2025bifold}
O.~Barbany, A.~Colom{\'e}, and C.~Torras, ``Bifold: Bimanual cloth folding with language guidance,'' \emph{arXiv preprint arXiv:2501.16458}, 2025.

\bibitem{obata2024lip}
K.~Obata, T.~Aoki, T.~Horii, T.~Taniguchi, and T.~Nagai, ``Lip-llm: Integrating linear programming and dependency graph with large language models for multi-robot task planning,'' \emph{IEEE Robotics and Automation Letters}, 2024.

\bibitem{jiang2025hiddendetect}
Y.~Jiang, X.~Gao, T.~Peng, Y.~Tan, X.~Zhu, B.~Zheng, and X.~Yue, ``Hiddendetect: Detecting jailbreak attacks against large vision-language models via monitoring hidden states,'' \emph{arXiv preprint arXiv:2502.14744}, 2025.

\bibitem{bi2024llava}
J.~Bi, Y.~Wang, H.~Chen, X.~Xiao, A.~Hecker, V.~Tresp, and Y.~Ma, ``Llava steering: Visual instruction tuning with 500x fewer parameters through modality linear representation-steering,'' \emph{arXiv preprint arXiv:2412.12359}, 2024.

\bibitem{diao2025temporal}
X.~Diao, C.~Zhang, W.~Wu, Z.~Ouyang, P.~Qing, M.~Cheng, S.~Vosoughi, and J.~Gui, ``Temporal working memory: Query-guided segment refinement for enhanced multimodal understanding,'' \emph{arXiv preprint arXiv:2502.06020}, 2025.

\bibitem{FineCIR}
Z.~Li, Z.~Fu, Y.~Hu, Z.~Chen, H.~Wen, and L.~Nie, ``Finecir: Explicit parsing of fine-grained modification semantics for composed image retrieval,'' \emph{https://arxiv.org/abs/2503.21309}, 2025.

\bibitem{tschannen2025siglip}
M.~Tschannen, A.~Gritsenko, X.~Wang, M.~F. Naeem, I.~Alabdulmohsin, N.~Parthasarathy, T.~Evans, L.~Beyer, Y.~Xia, B.~Mustafa \emph{et~al.}, ``Siglip 2: Multilingual vision-language encoders with improved semantic understanding, localization, and dense features,'' \emph{arXiv preprint arXiv:2502.14786}, 2025.

\bibitem{liu2024dora}
S.-Y. Liu, C.-Y. Wang, H.~Yin, P.~Molchanov, Y.-C.~F. Wang, K.-T. Cheng, and M.-H. Chen, ``Dora: Weight-decomposed low-rank adaptation,'' in \emph{Forty-first International Conference on Machine Learning}, 2024.

\bibitem{xin2024mmap}
Y.~Xin, J.~Du, Q.~Wang, K.~Yan, and S.~Ding, ``Mmap: Multi-modal alignment prompt for cross-domain multi-task learning,'' in \emph{Proceedings of the AAAI Conference on Artificial Intelligence}, vol.~38, no.~14, 2024, pp. 16\,076--16\,084.

\bibitem{zheng2024gaussiangrasper}
Y.~Zheng, X.~Chen, Y.~Zheng, S.~Gu, R.~Yang, B.~Jin, P.~Li, C.~Zhong, Z.~Wang, L.~Liu \emph{et~al.}, ``Gaussiangrasper: 3d language gaussian splatting for open-vocabulary robotic grasping,'' \emph{IEEE Robotics and Automation Letters}, 2024.

\bibitem{wen2025tinyvla}
J.~Wen, Y.~Zhu, J.~Li, M.~Zhu, Z.~Tang, K.~Wu, Z.~Xu, N.~Liu, R.~Cheng, C.~Shen \emph{et~al.}, ``Tinyvla: Towards fast, data-efficient vision-language-action models for robotic manipulation,'' \emph{IEEE Robotics and Automation Letters}, 2025.

\bibitem{xu2024naturalvlm}
R.~Xu, Y.~Shen, X.~Li, R.~Wu, and H.~Dong, ``Naturalvlm: Leveraging fine-grained natural language for affordance-guided visual manipulation,'' \emph{IEEE Robotics and Automation Letters}, 2024.

\bibitem{jiang2025screencoder}
Y.~Jiang, Y.~Zheng, Y.~Wan, J.~Han, Q.~Wang, M.~R. Lyu, and X.~Yue, ``Screencoder: Advancing visual-to-code generation for front-end automation via modular multimodal agents,'' \emph{arXiv preprint arXiv:2507.22827}, 2025.

\bibitem{bi2025prism}
J.~Bi, Y.~Wang, D.~Yan, X.~Xiao, A.~Hecker, V.~Tresp, and Y.~Ma, ``Prism: Self-pruning intrinsic selection method for training-free multimodal data selection,'' \emph{arXiv preprint arXiv:2502.12119}, 2025.

\bibitem{zhao2025learning}
H.~Zhao, J.~Zhu, Z.~Yan, Y.~Li, Y.~Deng, and X.~Wang, ``Learning generalizable language-conditioned cloth manipulation from long demonstrations,'' \emph{arXiv preprint arXiv:2503.04557}, 2025.

\bibitem{li2025language}
Z.~Li, J.~Liu, Z.~Li, Z.~Dong, T.~Teng, Y.~Ou, D.~Caldwell, and F.~Chen, ``Language-guided dexterous functional grasping by llm generated grasp functionality and synergy for humanoid manipulation,'' \emph{IEEE Transactions on Automation Science and Engineering}, 2025.

\bibitem{wei2022chain}
J.~Wei, X.~Wang, D.~Schuurmans, M.~Bosma, F.~Xia, E.~Chi, Q.~V. Le, D.~Zhou \emph{et~al.}, ``Chain-of-thought prompting elicits reasoning in large language models,'' \emph{Advances in neural information processing systems}, vol.~35, pp. 24\,824--24\,837, 2022.

\bibitem{lin2023text2motion}
K.~Lin, C.~Agia, T.~Migimatsu, M.~Pavone, and J.~Bohg, ``Text2motion: From natural language instructions to feasible plans,'' \emph{Autonomous Robots}, vol.~47, no.~8, pp. 1345--1365, 2023.

\bibitem{wang2023learning}
X.~Wang, W.~Wang, J.~Shao, and Y.~Yang, ``Learning to follow and generate instructions for language-capable navigation,'' \emph{IEEE Transactions on Pattern Analysis and Machine Intelligence}, vol.~46, no.~5, pp. 3334--3350, 2023.

\bibitem{huang2022mesh}
Z.~Huang, X.~Lin, and D.~Held, ``Mesh-based dynamics with occlusion reasoning for cloth manipulation,'' in \emph{Robotics: Science and Systems (RSS)}, 2022.

\bibitem{10411033}
D.~Zheng, S.~Yao, W.~Xu, and C.~Lu, ``Differentiable cloth parameter identification and state estimation in manipulation,'' \emph{IEEE Robotics and Automation Letters}, vol.~9, no.~3, pp. 2519--2526, 2024.

\bibitem{zhou2024imitating}
P.~Zhou, J.~Qi, A.~Duan, S.~Huo, Z.~Wu, and D.~Navarro-Alarcon, ``Imitating tool-based garment folding from a single visual observation using hand-object graph dynamics,'' \emph{IEEE Transactions on Industrial Informatics}, vol.~20, no.~4, pp. 6245--6256, 2024.

\bibitem{doumanoglou2016folding}
A.~Doumanoglou, J.~Stria, G.~Peleka, I.~Mariolis, V.~Petrik, A.~Kargakos, L.~Wagner, V.~Hlav{\'a}{\v{c}}, T.-K. Kim, and S.~Malassiotis, ``Folding clothes autonomously: A complete pipeline,'' \emph{IEEE Transactions on Robotics}, vol.~32, no.~6, pp. 1461--1478, 2016.

\bibitem{wang2024rl}
Y.~Wang, Z.~Sun, J.~Zhang, Z.~Xian, E.~Biyik, D.~Held, and Z.~Erickson, ``Rl-vlm-f: reinforcement learning from vision language foundation model feedback,'' in \emph{Proceedings of the 41st International Conference on Machine Learning}, 2024, pp. 51\,484--51\,501.

\bibitem{tian2025diffusion}
T.~Tian, H.~Li, B.~Ai, X.~Yuan, Z.~Huang, and H.~Su, ``Diffusion dynamics models with generative state estimation for cloth manipulation,'' \emph{arXiv preprint arXiv:2503.11999}, 2025.

\bibitem{chen2025metafold}
H.~Chen, J.~Li, R.~Wu, Y.~Liu, Y.~Hou, Z.~Xu, J.~Guo, C.~Gao, Z.~Wei, S.~Xu \emph{et~al.}, ``Metafold: Language-guided multi-category garment folding framework via trajectory generation and foundation model,'' \emph{arXiv preprint arXiv:2503.08372}, 2025.

\bibitem{liu2024grounding}
S.~Liu, Z.~Zeng, T.~Ren, F.~Li, H.~Zhang, J.~Yang, Q.~Jiang, C.~Li, J.~Yang, H.~Su \emph{et~al.}, ``Grounding dino: Marrying dino with grounded pre-training for open-set object detection,'' in \emph{European conference on computer vision}.\hskip 1em plus 0.5em minus 0.4em\relax Springer, 2024, pp. 38--55.

\bibitem{kirillov2023segment}
A.~Kirillov, E.~Mintun, N.~Ravi, H.~Mao, C.~Rolland, L.~Gustafson, T.~Xiao, S.~Whitehead, A.~C. Berg, W.-Y. Lo \emph{et~al.}, ``Segment anything,'' in \emph{Proceedings of the IEEE/CVF international conference on computer vision}, 2023, pp. 4015--4026.

\bibitem{li2023blip}
J.~Li, D.~Li, S.~Savarese, and S.~Hoi, ``Blip-2: Bootstrapping language-image pre-training with frozen image encoders and large language models,'' in \emph{International conference on machine learning}.\hskip 1em plus 0.5em minus 0.4em\relax PMLR, 2023, pp. 19\,730--19\,742.

\bibitem{radford2021learning}
A.~Radford, J.~W. Kim, C.~Hallacy, A.~Ramesh, G.~Goh, S.~Agarwal, G.~Sastry, A.~Askell, P.~Mishkin, J.~Clark \emph{et~al.}, ``Learning transferable visual models from natural language supervision,'' in \emph{International conference on machine learning}.\hskip 1em plus 0.5em minus 0.4em\relax PmLR, 2021, pp. 8748--8763.

\bibitem{lin2021softgym}
X.~Lin, Y.~Wang, J.~Olkin, and D.~Held, ``Softgym: Benchmarking deep reinforcement learning for deformable object manipulation,'' in \emph{Conference on Robot Learning}.\hskip 1em plus 0.5em minus 0.4em\relax PMLR, 2021, pp. 432--448.

\bibitem{hu2022lora}
E.~J. Hu, yelong shen, P.~Wallis, Z.~Allen-Zhu, Y.~Li, S.~Wang, L.~Wang, and W.~Chen, ``Lo{RA}: Low-rank adaptation of large language models,'' in \emph{International Conference on Learning Representations}, 2022.

\bibitem{liu2022few}
H.~Liu, D.~Tam, M.~Muqeeth, J.~Mohta, T.~Huang, M.~Bansal, and C.~A. Raffel, ``Few-shot parameter-efficient fine-tuning is better and cheaper than in-context learning,'' \emph{Advances in Neural Information Processing Systems}, vol.~35, pp. 1950--1965, 2022.

\bibitem{team2025doubao}
D.~Team, ``Doubao-1.5-pro,'' 2025.

\bibitem{xAIGrok3_2025}
xAI, ``Grok 3: Advancing real-time reasoning in {{AI}},'' Tech. Rep., 2025.

\bibitem{liu2024deepseek}
A.~Liu, B.~Feng, B.~Xue, B.~Wang, B.~Wu, C.~Lu, C.~Zhao, C.~Deng, C.~Zhang, C.~Ruan \emph{et~al.}, ``Deepseek-v3 technical report,'' \emph{arXiv preprint arXiv:2412.19437}, 2024.

\bibitem{hurst2024gpt}
A.~Hurst, A.~Lerer, A.~P. Goucher, A.~Perelman, A.~Ramesh, A.~Clark, A.~Ostrow, A.~Welihinda, A.~Hayes, A.~Radford \emph{et~al.}, ``Gpt-4o system card,'' \emph{arXiv preprint arXiv:2410.21276}, 2024.

\end{thebibliography}
}

\end{document}